# Replica Exchange using q-Gaussian Swarm Quantum Particle Intelligence Method

Hiqmet Kamberaj[1]

*Faculty of Technical Sciences, Department of Information Technology, International Balkan University, Skopje, R. of Macedonia*

Abstract

We present a newly developed **R**eplica **E**xchange algorithm using $q$-**G**aussian **S**warm **Q**uantum **P**article **O**ptimization (REX@q-GSQPO) method for solving the problem of finding the global optimum. The basis of the algorithm is to run multiple copies of independent swarms at different values of $q$ parameter. Based on an *energy* criterion, chosen to satisfy the *detailed balance*, we are swapping the particle coordinates of neighboring swarms at regular iteration intervals. The swarm replicas with high $q$ values are characterized by high diversity of particles allowing escaping local minima faster, while the low $q$ replicas, characterized by low diversity of particles, are used to sample more efficiently the local basins. We compare the new algorithm with the standard Gaussian Swarm Quantum Particle Optimization (GSQPO) and q-Gaussian Swarm Quantum Particle Optimization (q-GSQPO) algorithms, and we found that the new algorithm is more robust in terms of the number of fitness function calls, and more efficient in terms ability convergence to the global minimum. In additional, we also provide a method of optimally allocating the swarm replicas among different $q$ values. Our algorithm is tested for three benchmark functions, which are known to be multimodal problems, at different dimensionalities. In addition, we considered a polyalanine peptide of 12 residues modeled using a Gō coarse-graining potential energy function.

**Introduction**

The problem of finding the global optimum in a multimodal and multidimensional space can be extremely difficult since the number of stable optima increases as the search space increases, for instance, the search for the global minimum energy in a surface energy landscape of the atomic structures. [1, 2] Swarm Particle Optimization (SPO) is population-based optimization technique, similar to evolutionary algorithms. [3] Kennedy & Eberhart introduced the method to solve the problem of finding the global optimum of a $d$ dimensional function. [4] The basis of SPO method are the swarm intelligence algorithms, which concern with the design of intelligent multi-agent systems based on the collective behavior of insects (ants, termites, bees, and wasps) or other animal societies (flocks of birds and schools of fish). [4]

In SPO method, the swarm particles, representing possible solutions, search the phase space, defined by their velocities and coordinates, which are updated based on the particle's own experience and experience of the particle's neighbors or the experience of the whole swarm. The method has already been used to solve many optimization problems, [5] with some interests also in other fields, such as statistical mechanics. [6]

Since the standard SPO algorithm has a low convergence rate, [7, 8] several improvements and variants of the SPO algorithm have been proposed. [9, 10, 11, 12, 13] The new variant of the SPO method, the so-called Swarm Quantum Particle Optimization (SQPO), has been considered as an improvement against the *classical* SPO method, since there is a nonzero probability to escape the local minima even for very high barriers. [14] Efforts have been made to improve the SQPO method. [15, 16, 17, 18, 19, 20, 21, 22] These improvements focus primarily on parameter selection criteria, [18, 22] and maintaining diversity of the swarm. [19, 20, 21] A detailed review of all these methods is described in Ref. [23] Use of different forms of attractive potential-energy surfaces for SQPO algorithm is also considered for improvement of the algorithm. [14] Different potentials yield different probability distributions, which describe the probability of finding the swarm quantum-like particle at a certain position in the phase space. [14]

---

[1] Corresponding author: hkamberaj@ibu.edu.mk; Tel. +389(0)75462189; Fax. +389(0)23214832; Address: International Balkan University, Tashko Karadza 11A, 1000 Skopje, R. of Macedonia

In our previous study [24], we showed that the use of q-Gaussian probability distribution of swarm particles (q-GSQPO algorithm) improved significantly the efficiency of searching for the global minimum when compared with Gaussian probability distribution of swarm particles (GSQPO algorithm). The q-Gaussian distribution is characterized by long tails, which allow for reaching long distant regions in phase space by increasing the diversity of the swarm particles. [24] The application of the probability distributions with heavy tails is found to be useful in allowing the system to escape from local optima in multimodal problems also in other areas [25, 26, 27, 28, 29], since the nonzero occupations taken from long tails of the distribution implies jumps of scale-free sizes. The $q$-Gaussian distribution is also known as Tsallis distribution [30] in statistical mechanics, and it is known to produce a smoothed potential energy surface. [31]

The problem facing GSQPO algorithm is the premature convergence to a local minima due to low diversity of the swarm particles. [32] This problem of GSQPO algorithm can efficiently be solved by using the q-GSQPO algorithm [24], which increases the diversity between the swarm particles, and hence allowing the swarm to explore long distant regions of the searching space.

In this paper we will investigate the use of the popular replica exchange method [33, 34], which is known for overcoming the problem of the sampling convergence in dynamical systems. In this method several copies of the system (so-called *replicas*) are run independently at different values of some internal parameter of the system (*e.g.*, temperature [33, 34], strength of interaction [35], and $q$ parameter when combined with Tsallis statistics [27, 36]). At regular time intervals the coordinates of the replicas are swapped based on an energy criterion that preserves detailed balance. The higher parameter value replicas are used to enhance the barrier crossings and the low parameter value replicas are used to sample the local basins.

Similarly, in this study, we will create $M$ copies of swarm particles (replicas) and run them independently at different values of $q$ where the lowest level replica is running at $q=1$. Then, in analogy with standard replica exchange methods, at regular iteration intervals we swap the particles of swarm replicas between two neighboring values of $q$ based on an *energy* criterion, which is explained in details in the next section. To increase the efficiency of the algorithm, we also discuss the optimization of the *round-trip time* of the swarm replicas between the lowest and highest values of $q$ and vice-versa. To test our algorithm, we studied three benchmark functions, namely the Ackley [37], Griewank [38] and Rastrigin [39] functions at different dimensions, $d = 5, 10, 20$ and $50$. In addition, we considered a peptide of 12 Alanine (Ala) residues using a Gō coarse-graining model for the potential energy function. [40]

**Materials and Methods**

**Theoretical basics of generalized q-GSQPO algorithm**

The swarm quantum particle optimization algorithm determines the probability of finding the swarm particle at the position $\vec{X}$ at any time $t$. [14] Details of the derivation of GSQPO and q-GSQPO algorithms can be found in our previous work. [24] Here, we re-write the equations, proposed to describe the q-GSQPO algorithm in a generalized, form as [24]:

$$\begin{cases} X_{ij}(t+1) = p_j + \gamma_t^{(q)} \left| M_j^{Best} - X_{ij}(t) \right| F_q(u) & z \geq 0.5 \\ X_{ij}(t+1) = p_j - \gamma_t^{(q)} \left| M_j^{Best} - X_{ij}(t) \right| F_q(u) & z < 0.5 \end{cases} \quad (1)$$

where $\gamma_t^{(q)}$ has to be less than 1.7 in order to guarantee the convergence of the particles, [24] and it is chosen here to have a simplified sinusoidal expression as

$$\gamma_t^{(q)} = 1 + g|A\sin(\omega t)| \tag{2}$$

with $g$ being a scaling constant fixed in this study to 0.5, $\omega = 0.1$, and $A$ was varying from 0.01 to 1. $\vec{M}^{Best}$, also called *Mainstream Thought* or *Mean Best* of the population [22], defines the mean of the $\vec{X}_i^{LBest}$ best positions of all particles ($i = 1, 2, \cdots, N$) in each dimension and it is given by [22]

$$\vec{M}^{Best} = \frac{1}{N}\sum_{i=1}^{N}\vec{X}_i^{LBest}. \tag{3}$$

Here, $F_q(u)$ is a random number following $q$-Gaussian distribution if $q > 1$ and Gaussian distribution if $q = 1$.

**Replica Exchange Method**

In the **R**eplica **E**xchange **q-GSQPO** (REXqQSPO) algorithm we replicate $M$ copies of the swarm particles among different values of $q$: $q_1, q_2, \ldots, q_M$. A geometrical distribution in the interval $[1, q_{max})$ was used to select the values of $q$, where $q_{max}$ is a maximum value of $q$, chosen here to be 3. In order to control the acceptance probability of swapping swarm particles between two neighboring $q$, we introduce a ``temperature'' parameter at each level:

$$\alpha_i = \frac{1}{kq_i} \tag{4}$$

where $i = 1, 2, \ldots, M$. $k$ is considered an adjustable parameter, which will be optimized to minimize the time of a round trip from the lowest to the highest value of $q$ and vice-versa. At regular iteration steps we swap the configurations of swarm particles between two randomly chosen neighboring $q$, let say $i$ and $j$, such that detailed balance is preserved:

$$P_i(q)T_{i \to j} = P_j(q)T_{j \to i} \tag{5}$$

$T_{i \to j}$ is the transition probability from the swarm replica $i$ to $j$, and $T_{j \to i}$, which is the transition probability from the swarm replica $j$ to $i$. We will assume that the probability of finding the swarm $i$ at a value $q$ is proportional to the Boltzmann factor as:

$$P_i(q) \propto \exp(-\alpha_i E_i)$$

where $E_i$ is the minimum of the best local scoring value of the swarm particles in replica $i$ at a given iteration step. Choosing $E_i$ as the minimum of the best local scoring value is optional. Other choices may also be allowed, for instance, the mean best local scoring value or the best global scoring value of the replica. The important is that the choice should yield a high value of the diversity of swarm particles on each of replica in order to allow the swarm particles to explore more efficiently the searching space. This is in particular important for low $q$ replicas, which are used to better explore the local basins.

Then, the probability of accepting an attempting swap between swarm replicas $i$ and $j$ is given by:

$$P_{acc}(i \leftrightarrow j) = \frac{T_{i \to j}}{T_{j \to i}} = \min\{1, \exp(-(\alpha_i - \alpha_j)(E_j - E_i))\} \tag{6}$$

The algorithm stops for searching the phase space when at least one of the swarm replicas converges to the optimal solution.

**Analyzing the results**

In order to estimate the efficiency of the algorithm, we calculated the frequencies of visiting each $q$ from each swarm replicas and compared the observed frequencies for each swarm with expected frequencies, which depend only on the number of replicas. We used the $\chi^2$ test with a confidence level of 95 % to investigate the goodness of the comparison. [41] In this study we chose a geometrical distribution of $q$ among 5 or 6 replicas. The replicated exchange simulations were stopped when the first global best score of the swarms was less than some minimum value, chosen here to be $10^{-5}$. As an estimation of the computational robustness of the method we used the number of iterations needed until the convergence is reached.

In addition, we also measure the highest and the lowest values of the diversity, which is given by [24]

$$\Delta = \frac{1}{N} \sum_{i=1}^{N} \left| \vec{X}_i^{LBest}(t) - \vec{M}^{Best}(t) \right| \tag{7}$$

For the polyalanine peptide, the **R**oot **M**ean **S**quare **D**eviation (rmsd) from a reference structure is calculated as:

$$rmsd(t) = \sqrt{\frac{1}{N} \sum_{i=1}^{N} \left| \mathbf{r}_i(t) - \mathbf{r}_{i,0}^{ref} \right|^2} \tag{8}$$

where $N$ is the number of residues and $\mathbf{r}_{i,0}^{ref}$ represents the vector of coordinates of the reference structure for the residue $i$.

**Results and Discussions**

**Test 1**

We considered the motion in three $d$ - dimensional potential functions determined mathematically as in Table I.

*Adjusting temperature parameter $k$*
Value of $k$ will influence on the average allocation of the swarm replicas over the $q$ space ladder, since the overlapping of probability distributions between two neighboring $q$ values, and hence the acceptance probability of swapping the swarm replicas, depend on $k$. Due to the barriers that exist in the $q$ space, the swarms at the low $q$ values may not be able to visit the upper values and vice-versa. In order to adjust the value of $k$ we did several run tests, where we compared the computed frequencies of observing the swarms at a value of $q$ with the expected frequencies assuming equal probability of finding a swarm at each value of $q$. In FIG. 1 we show the ratio of the computed $\chi^2$ with critical value of $\chi_c^2$ at confidence level of 95 % for the three benchmark functions at different values of parameter $k$. The dimension the searching space was fixed to $d = 10$. For $\chi^2 / \chi_c^2 < 1$, the probability distribution of swarm replicas in the $q$ space is considered to be uniform, at a confidence level of 95 %, otherwise there is not enough evidence to say that this distribution is uniform. We found that for values of $k < 1$, depending also on the benchmark function, there is enough evidence to say at the confidence level of 95 %, that the distribution of the swarm replicas in the $q$ space is uniform (see FIG. 1).

Table I. The three benchmark functions which are used in this study.

| Benchmark Function | Name and Reference |
|---|---|
| $f_1(x_1, x_2, \cdots, x_d) = \frac{1}{400}\sum_{i=1}^{d} x_i^2 - \prod_{i=1}^{d}\cos(x_i/\sqrt{i}) + 1,$ $-6\pi \leq x_i \leq 6\pi$ $f_1^*(x_i^*) = 0, \ x_i^* = 0, i = 1, 2, \cdots, d$ | Griewank [38] |
| $f_2(x_1, x_2, \cdots, x_d) = \sum_{i=1}^{d}\left(10 + x_i^2 - 10\cdot\cos(\pi x_i)\right),$ $-\pi/2 \leq x_i \leq \pi/2$ $f_2^*(x_i^*) = 0, \ x_i^* = 0, i = 1, 2, \cdots, d$ | Rastrigin [39] |
| $f_3(x_1, x_2, \cdots, x_d) = -20\exp\left(-0.2\sqrt{\frac{1}{d}\sum_{i=1}^{d} x_i^2}\right) -$ $exp\left(\frac{1}{d}\sum_{i=1}^{d}\cos(2\pi x_i)\right) + 20 + e,$ $-6\pi \leq x_i \leq 6\pi$ $f_3^*(x_i^*) = 0, \ x_i^* = 0, i = 1, 2, \cdots, d$ | Ackley [37] |

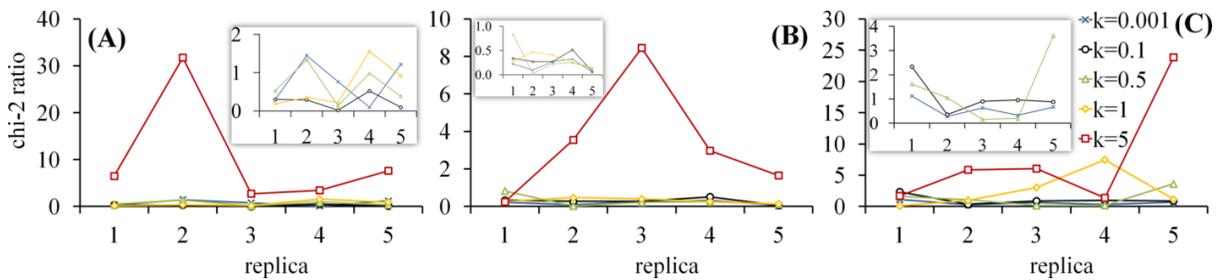

FIG. 1. The ratio between the calculated and critical value of $\chi^2$ at the confidence level of 95 % for different values of the parameter $k$. (A) For Ackley function; (B) Griewank function and (C) Rastrigin function. Dimensionality of the space was fixed to $d = 10$, and $q$ was chosen from a geometrical distribution in the range between one and 3.

*Robustness*
In order to examine the robustness of the REX@q-GSQPO algorithm in comparison with other algorithms such as GSQPO and q-GSQPO in terms of function calls, we studied the number of iterations needed for each algorithm until the convergence was reached for a fixed value of $k < 0.001$. For the REX@q-GSQPO algorithm we replicated five swarm particles among five values of $q$ in the range from 1 to 3 using a geometrical distribution. The swapping of the swarm particle configurations between two neighboring values of $q$ was performed at each iteration step. For the q-GSQPO algorithm we chose these values $q = 1.231, 1.414, 1.625, 1.866, 2.0$. The results of the number of iterations for the GSQPO ($q=1$) and q-GSQPO algorithms were averaged over different values of $A \in [0.01, 1]$.

Our results are presented graphically in FIG. 2 for the three benchmark functions versus the search space dimension $d$ (Ackley (A), Griewank (B) and Rastrigin (C)): in gray the average values for GSQPO and q-GSQPO algorithms, and in black for REX@q-GSQPO algorithm. Our data indicate that REX@q-GSQPO algorithm is more robust that GSQPO and q-GSQPO algorithms since the number of iterations needed for reaching the convergence is much smaller compare to GSQPO and q-GSQPO algorithms, in particular for high dimensionality which is of interest for many applications.

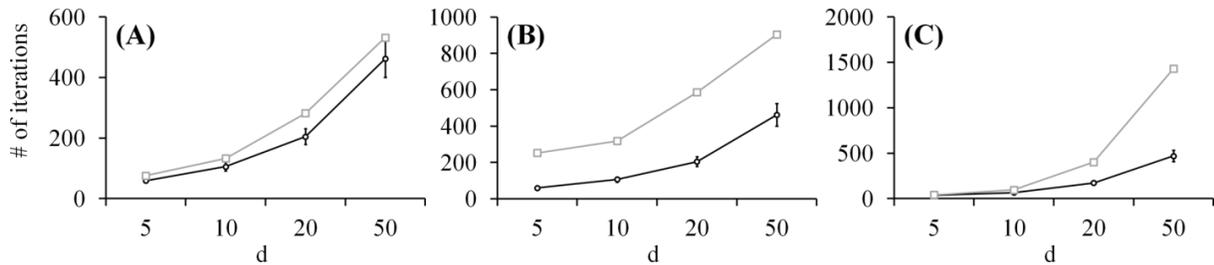

FIG. 2. The number of iterations as a function of the dimensionality of the problem $d = 5, 10, 20, 50$ for REX@q-GSQPO (black line) and combined GSQPO and q-GSQPO (gray line) algorithms. (A) Ackley function. (B) Griewank function. (C) Rastrigin function.

*Convergence*
To examine the efficiency of the method on finding the global minimum, we investigated the average best score and compared the three algorithms presented here. We used the same setup as described above. In FIG. 3 we are plotting the average best score for the REX@q-GSQPO algorithm (in black) and an average value for the GSQPO ($q=1$) and q-GSQPO ($q = 1.231, 1.414, 1.625, 1.866, 2.0$) algorithms (in gray). Results are presented for the three benchmark functions and different dimensions of the searching space $d$. It can be seen that REX@q-GSQPO algorithm provides much smaller best scoring values compare to standard GSQPO and q-GSQPO algorithms. In addition, for the REX@q-GSQPO algorithm in contrast to GSQPO and q-GSQPO algorithms, the convergence to the global minimum was achieved for all benchmark functions and for all dimensions considered here (see FIG. 3).

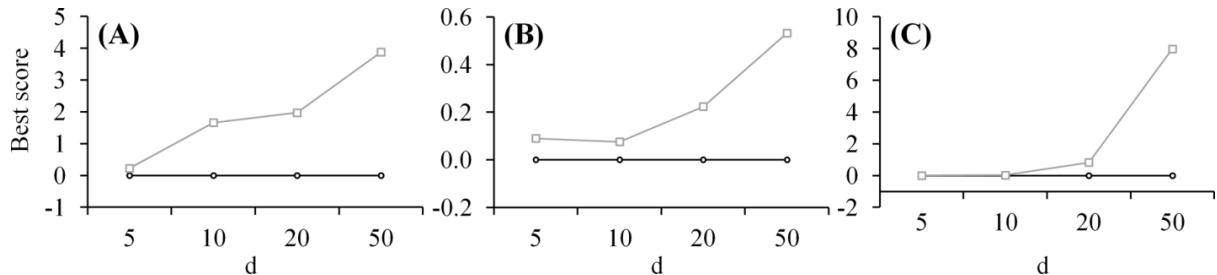

FIG. 3. The average best score value as a function of the dimensionality of the problem $d = 5, 10, 20, 50$ for REX@q-GSQPO (black line) and combined GSQPO and q-GSQPO (gray line) algorithms. (A) Ackley function. (B) Griewank function. (C) Rastrigin function.

**Test 2**

As a second test example, we considered a polyalanine peptide of 12 residues using the Gō coarse-graining model as depicted in FIG. 4.

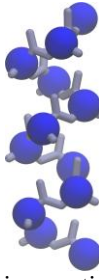

FIG. 4. The Gō coarse-graining model of the polyalanine peptide of 12 residues. Plot made using the VMD software. [43]

The potential energy function of the system is: [42]

$$E(r_1, r_2, \cdots, r_N) = \sum_{i=1}^{N-1}\left[k_{b,1}(d_{i,i+1}-d_0)^2 + k_{b,2}(d_{i,i+1}-d_0)^4\right]$$
$$+ \sum_{i=1}^{N-2} k_\theta (\theta_i - \theta_{0,i})^2$$
$$+ \sum_{i=1}^{N-3}\left[A(1+\cos\phi_i) + B(1+\cos 3\phi_i)\right] \quad (8)$$
$$+ \sum_{i<j}^{N_{nat}} 4\varepsilon\left[\left(\frac{\sigma_{ij}}{r_{ij}}\right)^{12} - \left(\frac{\sigma_{ij}}{r_{ij}}\right)^6\right]$$
$$+ \sum_{i<j}^{N_{nnat}} U_{ij}^{nnat}$$

with

$$U_{ij}^{nnat} = \begin{cases} 4\varepsilon\left[\left(\dfrac{\sigma_0}{r_{ij}}\right)^{12} - \left(\dfrac{\sigma_0}{r_{ij}}\right)^6\right] + \varepsilon, & r_{ij} < d_{cut} \\ 0, & r_{ij} > d_{cut} \end{cases}$$

The first term, in Eq. (8), represents bond stretching between two consecutive beads, where $d_{i,i+1}$ is the distance between the two consecutive beads, $d_0 = 3.8$ Å, $k_{l,1} = \varepsilon$ (kcal/Å$^2$), and $k_{l,2} = 100\varepsilon$ (kcal/Å$^4$). Here, $\varepsilon$ is the Lennard-Jones energy parameter chosen to be $\varepsilon = k_B T$ (kcal), where $k_B$ is the Boltzmann constant and $T = 300K$ is the temperature. The second term is the bond angle between three consecutive beads, where $k_\theta = 10\varepsilon$ (kcal/rad$^2$), and $\theta_{0,i}$ are equal to the bond angles in the native structure. The third term, in Eq. (8), is the dihedral angle energy between four consecutive beads, where $A = 0$ and $B = 0.2\varepsilon$ (kcal). The fourth term represents non-bounded interaction energy between two residues that are in contact in the native structure; the sum runs over all pairs of residues, $r_{ij}$ is the distance between two beads and $\sigma_{ij} = 2^{-1/6} d_{ij}$ with $d_{ij}$ being the corresponding distance in the native structure. The last term represents the potential of the non-native contacts, where $\sigma_0 = 2^{-1/6} d_{cut}$ and $d_{cut} = \langle d_{ij} \rangle$ which is the mean value of the lengths of the contacts. We started the simulations by uniformly distributing the residues in a box with each side in the range [-60, 60] and randomly assign them to each particle of the swarm and create in that way $M=6$ replicas distributed among $M=6$ values of $q$ chosen from a geometrical distribution in the interval [1, 3]. The dimensionality of the searching phase space is $d=3\times12=36$. We tried to swap the swarm particles between two neighboring replicas at the intervals of ten iterations. We calculated the rmsd from the reference structure, and plotted best score (kcal/mol), versus the rmsd as presented graphically in FIG. 5A. It can be seen that our simulations are able to distinguish five different minima at different significantly separated root mean square deviations from the reference structure. In addition, we show the highest and lowest diversities the swarms versus the number of iterations (see FIG. 5B). Our results indicate that high values of $q$ replicas are characterized by high diversity among the particles, as expected, while the low $q$ replicas are characterized by a low diversity. Hence, the high diversity swarm particles can be used to explore long distance regions of the searching space by allowing the particles to escape faster the local minima, and the low diversity swarm particles can be used to sample more efficiently the local basins.

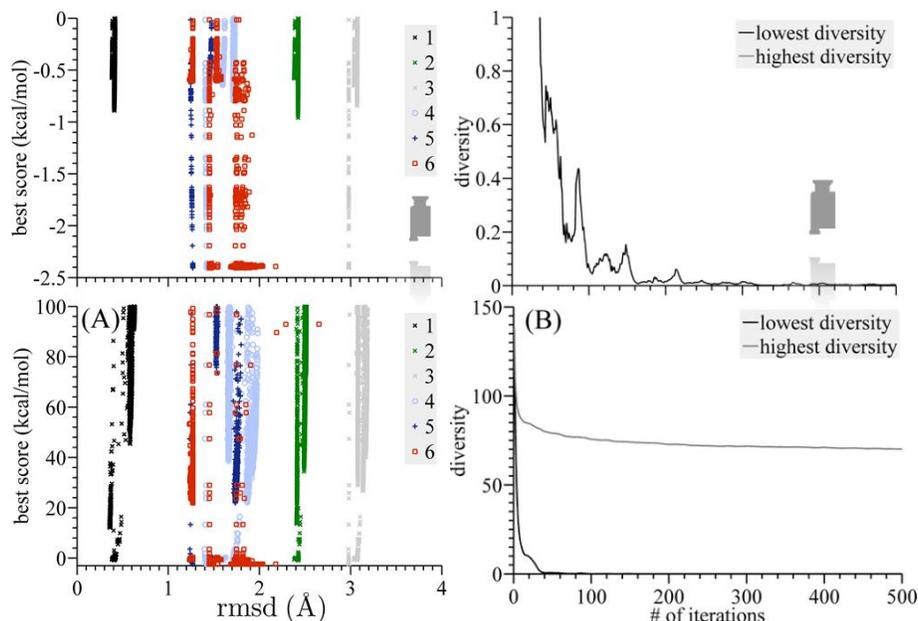

FIG. 5. (A) The best score potential energy (kcal/mol) versus the rmsd from the reference structure for each replica. (B) The highest and lowest value of the diversity among the replicas versus the number of iterations.

**Conclusions**

In this paper we presented a new algorithm for determining the global minimum of a multimodal problem using the replica exchange approach of the swarm quantum particles over the $q$ searching space. The algorithm was compared with standard Gaussian swarm quantum particles and q-Gaussian swarm quantum particles algorithms in terms of the efficiency of convergence to the global minimum and computational robustness.

It was found the new algorithm, REX@q-GSQPO, outperforms the standard algorithms GSQPO and q-GSQPO in both, the efficiency of the convergence and the computational robustness. In addition, we also examine and showed how to optimize the probability distribution among $q$ values in order to minimize the round-trip of the swarm replicas between the two extreme $q$ values. This was an important step to ensure the ergodicity in the $q$ space.

In the REX@q-GSQPO algorithm, the high $q$ values swarm replicas, which are characterized by high diversity between the particles, can be used to explore distant regions in the sampling space, while the low $q$ values, which are characterized by low diversity between the particles, can be used to sample efficiently the local basins.

Moreover, applying the algorithm for a polyalanine peptide, we showed that the method was able to distinguish between very close local minima in the energy space, but well separated in the rmsd space, making the method very suitable when applied to the problems of searching for the global minimum in rough potential energy landscapes.

## Acknowledgements

The author would like to thank the International Balkan University for the support.